\documentclass{ecai}

\usepackage{latexsym}
\usepackage{amssymb}
\usepackage{amsmath}
\usepackage{amsthm}
\usepackage{booktabs}
\usepackage{enumitem}
\usepackage{graphicx}
\usepackage{color}

\usepackage{url}

\usepackage[T1]{fontenc}
\usepackage[utf8]{inputenc}
\usepackage{microtype}
\usepackage{inconsolata}
\usepackage{relsize}
\usepackage{units}
\usepackage{hyperref}
\usepackage{caption}

\usepackage{multirow}
\usepackage{booktabs}
\usepackage{tabularx}
\newcolumntype{R}{>{\raggedleft\arraybackslash}X}
\newcolumntype{C}{>{\centering\arraybackslash}X}
\newcolumntype{L}{>{\raggedright\arraybackslash}X}

\usepackage{tcolorbox}
\usepackage{xcolor}
\newtcolorbox{chatboxDemoInput}[1][]{colback=gray!10, colframe=gray!60, sharp corners, boxrule=0.8pt, left=5pt, right=5pt, top=3pt, bottom=3pt, #1}
\newtcolorbox{chatboxDemoResponse}[1][]{colback=gray!10, colframe=gray!60, sharp corners, boxrule=0.8pt, left=5pt, right=5pt, top=3pt, bottom=3pt, #1}
\newtcolorbox{chatboxInput}[1][]{colback=blue!10, colframe=blue!60, sharp corners, boxrule=0.8pt, left=5pt, right=5pt, top=3pt, bottom=3pt, #1}
\newtcolorbox{chatboxResponse}[1][]{colback=blue!10, colframe=blue!60, sharp corners, boxrule=0.8pt, left=5pt, right=5pt, top=3pt, bottom=3pt, #1}

\usepackage[normalem]{ulem}
\newcommand\redsout{\bgroup\markoverwith{\textcolor{red}{\rule[0.5ex]{2pt}{0.4pt}}}\ULon}

\usepackage{acronym}

\acrodef{LLM}{large language model}
\acrodef{NLP}{natural language processing}
\acrodef{CoT}{chain-of-thought}
\acrodef{CoI}{chain-of-instructions}
\acrodef{SST}{Stanford Sentiment Treebank}
\acrodef{CR}{Customer-Reviews}
\acrodef{MR}{Movie-Reviews}
\acrodef{SQuAD}{Stanford Question Answering Dataset}
\acrodef{ASSET}{A Dataset of Sentence Simplification Evaluation Test}
\acrodef{SAMSum}{Summarizing Arguments in Online Discussions}
\acrodef{AGNews}{AG's News Topic Classification Dataset}
\acrodef{TREC}{Text REtrieval Conference}
\acrodef{DE}{Differential Evolution}
\acrodef{GA}{Genetic Algorithm}

\newcommand{\rqn}[1]{\texttt{RQ#1}}

\usepackage{pifont}
\newcommand{\cmark}{\ding{51}}%
\newcommand{\xmark}{\ding{55}}

\begin{document}

\begin{frontmatter}

    \paperid{5449}

    \title{A Toolbox for Improving Evolutionary Prompt Search}

    \author[A]{\fnms{Daniel}~\snm{Grie\ss{}haber}\footnote{Equal contribution.}\thanks{Corresponding Authors. E-mails: griesshaber@hdm-stuttgart.de, maximilian.kimmich@ims.uni-stuttgart.de.}}
    \author[B]{\fnms{Maximilian}~\snm{Kimmich}\footnotemark[1]\footnotemark[\ast]}
    \author[A]{\fnms{Johannes}~\snm{Maucher}}
    \author[B]{\fnms{Ngoc Thang}~\snm{Vu}}

    \address[A]{Institute for Applied Artificial Intelligence (IAAI), Stuttgart Media University}
    \address[B]{Institute for Natural Language Processing (IMS), University of Stuttgart}

    \begin{abstract}
        Evolutionary prompt optimization has demonstrated effectiveness in refining prompts for LLMs.
        However, existing approaches lack robust operators and efficient evaluation mechanisms.
        In this work, we propose several key improvements to evolutionary prompt optimization that can partially generalize to prompt optimization in general:
        1) decomposing evolution into distinct steps to enhance the evolution and its control, 2) introducing an LLM-based judge to verify the evolutions, 3) integrating human feedback to refine the evolutionary operator, and 4) developing more efficient evaluation strategies that maintain performance while reducing computational overhead.
        Our approach improves both optimization quality and efficiency.
        We release our code, enabling prompt optimization on new tasks and facilitating further research in this area.
    \end{abstract}

\end{frontmatter}


\section{Introduction}

The recent advent of \acp{LLM} has ushered in a new era of interactional artificial intelligence, democratizing access to powerful conversational agents and machine translation systems.
Despite impressive empirical gains, state-of-the‑art \acp{LLM} continue to exhibit notable gaps in genuine language understanding, often producing outputs that lack grounding.

One crucial bottleneck in harnessing \acp{LLM} for real‑world tasks lies in the formulation of \emph{prompts}: the textual instructions that guide model behavior.
Prompt quality has been shown to exert a profound influence on model performance, yet designing optimal prompts remains an art: manual tuning is labor‑intensive, ad hoc, and often fails to generalize across tasks or domains \citep{Brown2020,Liu2023}.
To overcome these limitations, a growing body of work has explored automatic prompt optimization methods, including evolutionary strategies in which candidate prompts are iteratively mutated and selected based on \ac{LLM} responses \citep{Shin2020,Deng2022,Shi2023,Guo2024}.
While promising, these approaches suffer from two key drawbacks: (i) their feedback loop relies on expensive API calls or compute resources to evaluate every candidate prompt, and (ii) their mutation operators themselves are typically hand‑crafted, limiting adaptability and often propagating hallucinations or other undesired artifacts \citep{Ji2023,Perkovic2024}.

One promising approach for improving prompt optimization is incorporating human feedback to verify and refine \ac{LLM} outputs \citep{Ouyang2022}.
While \acp{LLM} can automate prompt generation, human input remains crucial for verifying the accuracy of their results and correcting errors.
By integrating human feedback into the prompt optimization process, we aim to create a more robust system where humans not only verify \ac{LLM} output but also guide future prompt evolutions.
In the case that human feedback is not available (e.g., if there are no domain experts available, or it would be too costly), we allow another \ac{LLM} to act as judge and to take the responsibility of the human verifying the output of the evolutionary operator.
Additionally, since fewer instructions are simpler to verify and \ac{CoT} reasoning was shown to improve \ac{LLM} performance as well \citep{Wei2022}, we believe that the evolutionary operator as well as human feedback and the judging mechanism can benefit from more fine-grained instructions; we call this \ac{CoI}.
This design choice is motivated by several factors:
1) We hypothesize that reducing the complexity of individual instructions makes them easier for the model to follow, akin to how \ac{CoT} reasoning enhances model performance.
2) It minimizes confusion for the judge, allowing them to assess each step independently, rather than rejecting the entire evolutionary process due to errors in specific parts.
3) Human interaction becomes more efficient, as users can review and validate each evolution step separately -- similar to the judge -- and provide targeted feedback for improvements.
Furthermore, since the evaluation of generated prompts is crucial for the evolution, we introduce methods to efficiently evaluate prompts, finally leading to increased resource efficiency while maintaining task performance.

\begin{figure}[t]
    \centering
    \includegraphics[width=\columnwidth]{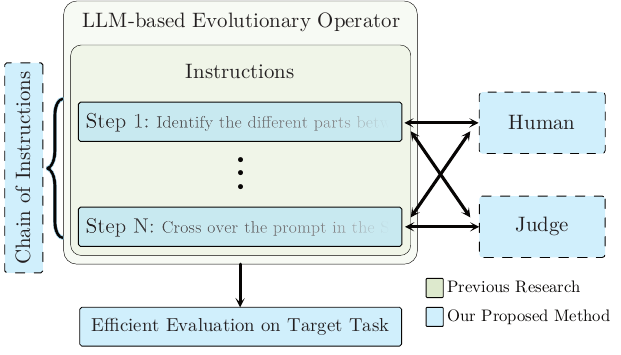}
    \caption{An overview of the individual components ascribed to our proposed method (blue) compared to only using a one-step instruction for the operator (green).}
    \vspace{10mm}
    \label{fig:method-overview}
\end{figure}

In this paper, we address the following research questions regarding \ac{LLM}-based evolutionary prompt optimization methods with the ultimate goal of improving language understanding in \acp{LLM}:
\begin{itemize}[noitemsep,topsep=0pt,align=left,leftmargin=*]
    \item[\rqn{1}:] How can new candidate prompts be evaluated more efficiently without forfeiting overall performance?
    \item[\rqn{2}:] Does \ac{CoI}-based evolution help judges and improve the evolutionary operator?
    \item[\rqn{3}:] Can \ac{LLM}-based judges assess the output quality of text-generative evolutionary operators?
    \item[\rqn{4}:] How can human feedback be effectively leveraged to enhance evolutionary prompt optimization?
    \item[\rqn{5}:] How does the selection of the \ac{LLM} influence the effectiveness of the proposed improvements?
\end{itemize}

To explore these questions, we propose an approach where the \ac{LLM} serves as both an operator for evolutions and a judge of prompts while allowing humans to intervene when the model output is incorrect.
This feedback loop allows human corrections to be leveraged 
for further optimizing the prompt search process.
Additionally, several evaluation strategies aim to make the evaluation more efficient and feedback loops faster.
An overview is given in figure \ref{fig:method-overview}.

Our contributions are as follows:
1) We introduce a novel human-in-the-loop approach for refining \ac{LLM}-based prompts, where the feedback is used for future optimizations.
2) We introduce an \ac{LLM}-based judge for verifying \ac{LLM}-based evolutions.
3) We present \acl{CoI} as a mechanism to enhance control over the evolutionary process and facilitate more effective feedback.
4) We show how the efficiency of prompt optimization methods can be increased without sacrificing performance.
5) We present empirical results investigating the benefits of our approaches in the context of evolutionary prompt optimization methods.
6) We publish all code that is necessary to reproduce the findings presented in this paper under a permissive license to enable further research into the topic of evolutionary prompt optimization and human feedback.

\section{Background}
In this section, we introduce the foundational concepts and methods that underpin our research on evolutionary prompt optimization.

\subsection{Large Language Models}
\acp{LLM} are a class of language models that leverage deep learning techniques to process and generate human language.
They have become increasingly popular recently due to their ability to generate high-quality text across a wide range of tasks \citep{Brown2020}.
\acp{LLM} are trained on large corpora of text data, enabling them to learn complex patterns and relationships within language.
They have been applied to a variety of \ac{NLP} tasks, including text generation, translation, summarization, and question answering.

The input for these \acp{LLM}, known as prompts, remains crucial since the LLMs' performance depends on these prompts.
The concept of \ac{CoT} reasoning has emerged as a powerful approach to enhancing the reasoning capabilities of \acp{LLM}.
\Ac{CoT} was first introduced by \citet{Wei2022}, demonstrating that prompting \acp{LLM} to decompose complex problems into intermediate reasoning steps significantly improves performance on several tasks.

In scientific and technical domains, the use of \acp{LLM} as judges has recently gained increasing attention.
Several studies have explored the potential of \acp{LLM} for assessing the quality of text in various contexts.
For instance, \cite{Zheng2023} investigated the ability of \acp{LLM} to grade academic writing and found that models like GPT-4 can provide feedback comparable to human reviewers.
Similarly, works by \citet{Liang2023} and \citet{Yu2024} have examined \acp{LLM} in the context of automated peer review, highlighting their strengths in identifying clarity issues and methodological flaws.

\subsection{Prompt Optimization}
Prompt optimization is the process of finding the most effective prompt for a given task.

For this work, prompt optimization approaches can be classified into two categories: continuous space optimization and discrete prompt optimization methods.
The former treat the prompt as a continuous vector, leveraging the fact that the tokens of the prompt are embedded into a continuous space, and optimize it using gradient-based techniques such as gradient descent \citep{Schick2021,Liu2022,Liu2023,Zhang2022}.
However, this approach requires the model to be differentiable with respect to the prompt, which may not always be the case.
Furthermore, the resulting prompt lacks interpretability by humans, as the resulting prompt is a continuous vector that cannot easily be mapped to discrete tokens, making it difficult to understand and refine.

Discrete prompt optimization methods, on the other hand, treat the prompt as a discrete sequence of tokens such that the result can easily be observed as a natural language prompt.
However, previous methods often rely on gradients over the model parameters to optimize the prompt, which may not be available in black-box models \citep{Shin2020,Deng2022,Shi2023,Zhang2023}.
In contrast, \citet{Guo2024} show that prompts can be optimized without gradients, enabling optimization methods for black-box models.

\subsection{Evolutionary Algorithms}
Evolutionary algorithms are a class of optimization techniques inspired by the process of natural selection.
They operate by iteratively evolving a population of candidate solutions to a problem, selecting the fittest individuals for reproduction and mutation.
Evolutionary algorithms have been successfully applied to a wide range of optimization problems, including function optimization, machine learning, and robotics \citep{Holland1992}.
\citet{Lehmann2022} demonstrated that \acp{LLM} can be used to perform automatic mutation of prompts, while \citet{Meyerson2024} showed the same for crossover operations, paving the way for evolutionary prompt optimization.
In the context of prompt optimization, evolutionary algorithms can be used to search for the best prompt for a given task by iteratively evolving and evaluating candidate prompts \citep{Chen2023}.
\citet{Guo2024} is the most related work to ours, as they use a genetic algorithm to optimize prompts for an \ac{LLM} for \ac{NLP} tasks.

\section{Method}
\label{sec:method}
Our goal is to find the ideal prompt $p$ for a given low-resource \ac{NLP} task (see section \ref{sec:training-tasks}) using in-context learning with an \ac{LLM}.
To achieve this goal, a small set of labeled validation data $\mathcal{D}$ is available ($\left|\mathcal{D}\right| \leq 200$).
The \ac{LLM} is treated as a black-box function, so no access to its parameters or inner architecture is available and needed, enabling models that are only available via APIs.

The optimization process resembles a genetic algorithm, where a population of prompts is evolved over $T$ generations.
\paragraph{Initialization:}
To generate the initial population, we select the best $\lfloor I/2\rfloor$ prompts from a task-specific set of base prompts and generate the remaining $\lceil I/2\rceil$ prompts by paraphrasing the selected base prompts, where $I$ is the population size.
This initialization warm-starts the optimization process with a diverse set of prompts \citep{So2019}.

\paragraph{Evolution:}
We follow the work of \citet{Lehmann2022,Meyerson2024,Guo2024} and use an \ac{LLM} 
to perform the operations of \textit{mutation} and \textit{crossover} of the evolutionary algorithm.
We use the same operator implementations for \ac{DE} and \ac{GA} as baselines to improve upon, also using demonstration data in the input for in-context learning.
That is, for the classification tasks, we randomly select one input-output pair per class, and a single example for the other tasks.

\paragraph{Evaluation:}
Prompt fitness $S_i$ is defined as the performance of an evaluation model with prompt $p_i$ on the validation data: $S_i = \mathcal{E}\left(p_i, \mathcal{D}\right)$.

\paragraph{Selection:} Prompts for evolution are selected following \citet{Lipowski2012} using a roulette-wheel style algorithm with stochastic sampling.
Specifically, a prompt is selected with probability $p_i = \frac{S_i}{\sum_{j=1}^{I} S_j}$.

\subsection{Efficient Evaluation}
The optimization process 
depends on the repeated evaluation of candidate prompts.
Although the evaluation set $\mathcal{D}$ is small, \ac{LLM} inference is still costly. 
Given the cost of a single inference step $c_i$, the total cost of the evaluation $c_e$ can be calculated as $c_e = c_i \times \left|\mathcal{D}\right| \times I \times T$.
Since $c_i$ is fixed to the model, a larger population size $I$ is generally understood to improve the result of the optimization, and the number of generations $T$ should be high enough to ensure convergence, we employ and present methods 
to reduce the overall cost of evaluation without negatively affecting the resulting prompt performance.

\subsubsection{Early Stopping}
The fitness function normally calculates a mean average score over all samples in $\mathcal{D}$.
Empirical preliminary examination has shown that the score converges before all samples have been tested.
Therefore, we evaluate strategies to reduce the number of evaluation inferences without affecting the resulting prompt performance.

\paragraph{Moment-based} We propose a moment-based early stopping strategy to stop the evaluation after the score has settled: If the mean absolute difference in evaluation score is less than a minimum change $\eta_m$ for a window size $w$, the evaluation is stopped.
The stopping criterion can be expressed with the following inequality:

\begin{equation*}
    \begin{split}
        \frac{1}{w} \sum_{n=t-w+1}^{t}\left|\mathcal{E}\left(p_i, \mathcal{D}_n\right) - \mathcal{E}\left(p_i, \mathcal{D}_{n-1}\right)\right| < \eta_m, \\t \geq w
    \end{split}
\end{equation*}

\paragraph{Parent-based} Except for the first generation, we have access to the performance of all ancestors of $p_i$ on samples from $\mathcal{D}$.
We propose to use this information for a parent-based early stopping decision to exit evaluation early if the current prompt $p_i$ is not performing better by $\eta_p$ compared to the max score of the parents ${p_a, p_b}$ in a sliding window $w$.
The stopping criterion is then fulfilled if the following inequality is true:

\begin{equation*}
    \begin{split}
        \max_{n = t - w + 1,\dots,t} \left(\mathcal{E}\left(p_i; \mathcal{D}_n\right) - \max\left( \mathcal{E}(p_a; \mathcal{D}_n),\, \mathcal{E}(p_b; \mathcal{D}_n) \right)\right) < \eta_p, \\t \geq w
    \end{split}
\end{equation*}

We employ the parent-based strategy for generations $T>1$ with a fallback to the moment-based strategy when parent performance is not available. Both early stopping methods employ a patience parameter to ignore the first evaluation iterations where the score may change drastically.

\subsubsection{Evaluation Strategies}
We consider different orderings of $\mathcal{D}$ for evaluation.
\paragraph{Shortest First}
With the motivation to reduce the number of tokens the \ac{LLM} needs to process during evaluation, the early stopping strategy can be extended to use an ordered version of $\mathcal{D}$ that is sorted in ascending order according to the length of the inputs.

\paragraph{Hardest First}
During evolution, the population of prompts is expected to improve.
We therefore propose an evaluation strategy that sorts the samples in $\mathcal{D}$ by the performance of the best parent prompt in ascending order.
This is motivated by the fact that a prompt that performs just as well as the best parent prompt on hard samples will not be able to yield a better mean performance on the whole dataset when including samples on which the parent already performed well.

\subsection{\acs{CoI} Prompting}
\label{sec:chain-of-instructions}
We adopt the concept of \ac{CoT} reasoning for our approach.
Rather than instructing the \ac{LLM} to reason step-by-step, we decompose the instructions for implementing the evolutionary operator into multiple distinct steps.
That is, for evolution step $t$, we formulate the prompt $o_t$ to include instructions $i_t$ and model response $r_t$ for previous steps as well as the instruction for the current step, $o_t=i_0,r_0,\ldots,i_{t-1},r_{t-1},i_t$.
Here, each instruction $i_t$ is a single operation that the \ac{LLM} should perform, such as mutating a prompting or crossing over two prompts.

When utilizing demonstration data, we ensure that it aligns with the current stage of evolution, meaning that instructions and model responses up to the current evolution step are included.

\subsection{Evolution Judge}
\label{sec:method-judge}
To avoid an expensive evaluation for prompts that are unlikely to be selected in absence of human feedback, we introduce a judge model $\mathcal{J}$.
For this, we use another \ac{LLM} to assess the quality of a prompt candidate $p_i$ before starting evaluation.
To this end, we provide the judge model with the response itself, along with the corresponding inputs that led to it -- including demonstration samples, system message and the prompt.
In case of \ac{CoI}, we apply the judge for each evolution step.
If the judge model determines a prompt to be of low quality, we ask the evolution model to generate a new response until a predefined number of repetitions is reached.
Afterward, if, according to the judge model, there is no prompt of high quality, we 
continue with a random response from the model.\footnote{In our implementation, we use the last one since there is no implication on the order of generated model responses, i.e., the randomness is realized via the evolution model.}

\subsection{Human Feedback}
\label{sec:human-feedback}
To integrate human feedback into the optimization process, we propose a human-in-the-loop approach that actively involves humans at multiple stages of evolutionary prompt optimization.
In our framework, human participants are not merely passive evaluators but play an active role in observing, analyzing, and refining the outputs generated by the \ac{LLM} during each step of the evolutionary process.

Specifically, after each evolutionary step -- such as mutation or crossover -- humans review the generated model outputs.
If deficiencies, ambiguities, or errors are detected, humans intervene by refining the instructions that guide the evolutionary operator.
This may include clarifying the language of the instructions, specifying more granular or explicit requirements, or restructuring the sequence of steps to reduce confusion for the \ac{LLM}.
Additionally, humans can update or augment the demonstration samples used for in-context learning, ensuring that these examples better illustrate the intended behavior and address previously observed shortcomings.

This process is inherently iterative: after each round of human intervention, the evolutionary process resumes with the updated instructions and demonstration data, allowing for continuous improvement.
Over successive cycles, this feedback loop enables the identification and mitigation of persistent weaknesses, such as the \ac{LLM}'s tendency to overlook subtle distinctions or to generate extraneous output, thereby enhancing the language understanding capabilities of the \ac{LLM}.
By systematically addressing these issues, the overall effectiveness and reliability of the prompt optimization process are enhanced.

Furthermore, this approach allows for the accumulation of best practices and refined instructions, which can be reused or adapted for future tasks or models.
The iterative nature of human feedback ensures that the optimization process remains adaptable and responsive to the evolving capabilities and limitations of the underlying \ac{LLM}.
Illustrative examples of how human feedback leads to tangible improvements in prompt optimization are provided in section \ref{sec:analysis-human}.

\section{Experimental Setup}
This section describes the experiments conducted to explore the effectiveness of the proposed methods. All code, data and information that is neccessary to reproduce the results of the presented experiments is published online.\footnote{
    \url{https://gitlab.mi.hdm-stuttgart.de/griesshaber/evoprompt}
}

\subsection{Tasks} \label{sec:training-tasks}
We evaluate our proposed method on a wide range of \ac{NLP} tasks, including sentiment analysis (\ac{SST} 2 \& 5 \citep{Socher2013}, \ac{MR} \citep{Maas2011}, \ac{CR} \citep{Hu2004}), subjectivity analysis (Subj \citep{Pang2002}), topic classification (\ac{AGNews} \citep{Zhang2015}, \ac{TREC} \citep{Li2002}), question answering (\ac{SQuAD} \citep{Rajpurkar2016}), simplification (\ac{ASSET} \citep{Alva2020}) and summarization (\ac{SAMSum} \citep{Gliwa2019}).

\noindent For evaluation of prompt fitness during evolution, we select a subset $\mathcal{D}$ of $200$ labeled samples from the validation set.
For the final evaluation, we use the whole test set to assess the performance of the evolved prompt.

\subsection{Evaluation Strategies}
To assess the effectiveness of the proposed evaluation strategies, the methods are compared to a baseline where prompt evaluation is performed on the whole validation set $\mathcal{D}$ and an additional na\"i{}ve strategy to reduce the evaluation cost by subsampling $\mathcal{D}$ with a fixed factor.
Since the fitness score of a prompt is an important metric in the evolutionary algorithm, we also show the score of the final prompt to preclude negative impacts on the final performance of the optimized prompt.

\subsection{\acs{CoI} Prompting}
We decompose the instructions of the evolutionary operators, DE and GA, into multiple steps as described in section \ref{sec:chain-of-instructions}.
Given the increased number of instructions required for \ac{DE}, the \ac{CoI}-based implementation consists of four steps, whereas \ac{GA} follows a more concise two-step process.
The baseline model for the \ac{CoI} experiments only performs a single step for the evolution of a prompt.
To see the combined effect of \ac{CoI} and judging, experiments using \ac{CoI} are performed with and without a judge $\mathcal{J}$.
For the baseline, the judge can only decide on the single output of the evolution.
Experiments are performed on and averaged over all tasks.

\subsection{Evolution Judge}
We employ a judge in our experiments to verify model responses in absence of humans, as described in section \ref{sec:method-judge}.
In detail, the LLM's instruction is given as follows: "You are acting as a judge. Please read the context, the instruction and the response and decide if the response follows the instruction. If it does, answer 'good'. If it does not, answer 'bad'. Wrap the answer with tags <judgement> and </judgement>. Please also add an explanation for your judgement."\footnote{This prompt is the result of a manual investigation; since this work deals with the optimization of prompts and the manual tuning of prompts contradicts the motivation of this work, this manual tuning is only necessary at this point, but neither reflects our intention to improve prompt optimization, nor does it counteract it.}.
We repeat generating model responses up to three times (if the judge assesses an ouput as bad).

Similar to the \ac{CoI} results, the experiments for the judge are compared with and without \ac{CoI} to assess the combined effect on the final performance.

\subsection{Human Feedback}
After analyzing the evolution model outputs, we iteratively refined the instructions for the evolutionary operator, re-evaluating and further improving the model outputs as needed.
Through this process, we applied two consecutive refinements to DE, resulting in $\text{DE}_1$ and $\text{DE}_2$.
Similarly, GA was improved once ($\text{GA}_1$), as its fewer instructions needed less care.
An example of such refinement is provided in section \ref{sec:analysis-human}.

\subsection{Hyperparameters}
For the evolution, we utilized the quantized version of Llama 3.1 8B Instruct \citep{Grattafiori2024} as the generative model.
The results are reported after conducting $T = 10$ generations with a population size of $I = 10$.
We adopt the approach of \citet{Guo2024} for selecting base prompts for the initial population described in section \ref{sec:method}.
Specifically, depending on the task, we utilize prompts from \citet{Mishra2021}, \citet{Zhang2022b}, \citet{Li2023}, \citet{Sanh2021}, \citet{Zhang2023b}.
However, for \ac{SQuAD}, we employ a single manually crafted prompt alongside generated prompts obtained using the forward mode generation method proposed by \citet{Zhou2022}.

For both paraphrasing and evolutionary steps, sampling was applied in decoding using a temperature of $t = 0.5$ to increase output variance.
We used the same model for the judge and for the evaluation, i.e., Llama 3.1 8B Instruct, but with greedy decoding for increased correctness.

For the early stopping, we set $\eta_m=10^{-3}$, the window size $w = 10$, $\eta_p = 10^{-3}$ and a patience of $20$.

\subsection{Other \acp{LLM}}
\label{sec:experimental-setup-other-models}

\makeatletter
\def\blfootnote{\gdef\@thefnmark{}\@footnotetext}
\makeatother

To explore the effect the choice of \ac{LLM} has on the performance of the proposed methods, we conduct experiments with numerous other \acp{LLM}\blfootnote{Model references on Hugging Face:} :
\begin{itemize}[align=left]
    \item[\texttt{LI3.1 (8B)}:] Llama 3.1 8B Instruct\footnote{meta-llama/Meta-Llama-3.1-8B-Instruct} (our base model).
    \item[\texttt{MI (7B)}:] Mistral Instruct 7B\footnote{mistralai/Mistral-7B-Instruct-v0.1}
    \item[\texttt{Q2.5 (7B)}:] Qwen 2.5 7B Instruct\footnote{Qwen/Qwen2.5-7B-Instruct}
    \item[\texttt{R1Q (1.5B)}:] Deep Seek R1 Distill Qwen 1.5B Instruct\footnote{deepseek-ai/DeepSeek-R1-Distill-Qwen-1.5B}
    \item[\texttt{R1Q (7B)}:] Deep Seek R1 Distill Qwen 7B Instruct\footnote{deepseek-ai/DeepSeek-R1-Distill-Qwen-7B}
    \item[\texttt{GI (1B)}:] Google Gemma 1B Instruct\footnote{google/gemma-1b-it}
    \item[\texttt{GI (7B)}:] Google Gemini 7B Instruct\footnote{google/gemma-7b-it}
\end{itemize}

\begin{table}[t!]
    \caption{%
        Baseline results for our hyperparameters and models for various classification, question answering, and text generation tasks.
        Results are reported for two evolutionary algorithms:
        Differential Evolution (DE) and Genetic Algorithm (GA).
        GA outperforms DE on most tasks, including AGNews, ASSET, SAMSum, SQuAD, SST-2, SST-5, and Subj.
        DE achieves better results on CR, MR, and TREC.
        This indicates that -- without any of our improvements -- while GA generally yields better overall performance, DE remains competitive on certain tasks.
    }
    \label{tab:results-baseline}
    \begin{tabularx}{\columnwidth}{lRR}
        \toprule
                    & \multicolumn{2}{r}{Evolution Algorithm}                     \\
        Task        & DE                                      & GA                \\
        \midrule
        \ac{AGNews} & 86.89\%                                 & \bfseries 87.99\% \\
        \ac{ASSET}  & 54.74\%                                 & \bfseries 56.78\% \\
        \ac{CR}     & \bfseries 91.85\%                       & 91.40\%           \\
        \ac{MR}     & \bfseries 91.30\%                       & 90.55\%           \\
        \ac{SAMSum} & 29.55\%                                 & \bfseries 29.82\% \\
        \ac{SQuAD}  & 86.72\%                                 & \bfseries 89.27\% \\
        \ac{SST} 2  & 94.51\%                                 & \bfseries 95.50\% \\
        \ac{SST} 5  & 56.24\%                                 & \bfseries 56.33\% \\
        Subj        & 80.75\%                                 & \bfseries 84.10\% \\
        \ac{TREC}   & \bfseries 83.20\%                       & 78.80\%           \\
        \bottomrule
    \end{tabularx}
\end{table}

\section{Quantitative Results}

This section presents the results of the conducted experiments in a quantitative manner to show the effectiveness of each modification, analyzing their implications with respect to our research questions.

\subsection{Baseline}

Since we aim to investigate the impact of our proposed extensions to evolutionary prompt optimization, table \ref{tab:results-baseline} presents baseline results for our chosen hyperparameters and \ac{LLM} models for all presented tasks and both evolution algorithms.
The results mostly replicate the final results presented in \citet{Guo2024}, but with deviations in hyperparameters and implementation.

\begin{table}[t]
    \caption{Comparing the different evaluation strategies to the baseline:
        average difference of evaluation scores ($\Delta S$), difference in number of tokens used during prompt evaluation ($\Delta c_e$), 
        token count ($\nicefrac{c_e}{c_b}$) and runtime ($\nicefrac{t_e}{t_b}$) of the experiment as a fraction of the baseline.
        All reported values are averaged over tasks.}
    \label{tab:results-strategy}
    \begin{tabularx}{\columnwidth}{Lrrrr}
        \toprule
        Strategy       & $\Delta S$        & $\Delta c_e$    & $\nicefrac{c_e}{c_b}$ & $\nicefrac{t_e}{t_b}$ \\
        \midrule
        Subsample      & -2.28\%           & -7.3M           & 33.5\%                & 55.6\%                \\
        \cline{1-5}
        Early Stopping & -1.43\%           & -7.6M           & 31.0\%                & 53.2\%                \\
        \cline{1-5}
        Shortest First & \bfseries +0.11\% & -7.9M           & 28.3\%                & 43.6\%                \\
        \cline{1-5}
        Hardest First  & -0.50\%           & \bfseries -8.0M & \bfseries 25.5\%      & \bfseries 42.8\%      \\
        \bottomrule
    \end{tabularx}
\end{table}

\subsection{\rqn{1}: Efficient Evaluation}

Table \ref{tab:results-strategy} shows the results for the experimental evaluation of the proposed efficient evaluation strategies.
The results indicate that, while all evaluation strategies reduce the number of tokens needed to score the candidate prompts, the na\"ive approach of subsampling $\mathcal{D}$ performs in average worst, 
demonstrating the importance of high-quality prompt candidate scores for the evolutionary algorithm.
Notably, the suggested strategies of evaluating on the shortest and hardest samples first only show minor deviation from the baseline scores while also being most effective reducing used tokens and runtime, whereas early stopping on the unordered scoring set may decrease performance. 
While the \textit{Hardest First} strategy reduces the evaluation cost the most, \textit{Shortest First} is the only strategy that did not show any decrease in the final evaluation.
In total, to answer \rqn{1}, both strategies, Hardest First and Shortest First, can effectively reduce the compute usage.
They should be chosen based on individual preferences on the final task performance.

\subsection{\rqn{2\&3}: \acs{CoI} Prompting \& Evolution Judge}

In table \ref{tab:results-coi}, the relative improvements that can be achieved with \ac{CoI} prompting are presented.
The average across tasks is positive for all configurations, regardless of evolution algorithm and the use of an additional judge.

Notably, the mean improvement for the \ac{DE} algorithm is higher than for the \ac{GA} algorithm, independent of whether the judge is used. Since \ac{DE} is more complex with a higher number of steps, this indicates that \ac{CoI} helps by breaking the algorithm into discrete steps that can be performed individually.
Furthermore, \ac{CoI} yields the best performance in combination with the judge, additionally motivating to verify model outputs automatically in \ac{LLM}-based evolutionary operators.

\begin{table}[t]
    \caption{%
        Performance improvements of \ac{CoI} over baselines with (\cmark) and without (\xmark) using a judge assessing model outputs.
        Incorporating \ac{CoI} consistently improves results across both Differential Evolution (DE) and Genetic Algorithm (GA), with a judge yielding best results for both algorithms.
        For DE, the mean improvement increases from +1.20\% to +1.73\%, with a higher maximum gain.
        Similarly, GA sees an increase in mean improvement from +0.68\% to +1.00\%, with a notable maximum gain of +4.40\%.
        These results demonstrate that \ac{CoI} optimization yields more reliable and higher-quality model outputs.
    }
    \label{tab:results-coi}
    \begin{tabularx}{\columnwidth}{ccRRR}
        \toprule
        Evolution              &               & \multicolumn{3}{c}{$\Delta S$}                               \\
        Algorithm              & $\mathcal{J}$ & \textit{mean}                  & \textit{min} & \textit{max} \\
        \midrule
        \multirow[c]{2}{*}{DE} & \xmark        & +1.20\%                        & -0.35\%      & +3.00\%      \\
        \cline{2-5}
                               & \cmark        & \bfseries +1.73\%              & -0.80\%      & +3.56\%      \\
        \cline{1-5} \cline{2-5}
        \multirow[c]{2}{*}{GA} & \xmark        & +0.68\%                        & -0.88\%      & +2.55\%      \\
        \cline{2-5}
                               & \cmark        & \bfseries +1.00\%              & -0.63\%      & +4.40\%      \\
        \bottomrule
    \end{tabularx}
\end{table}

\begin{table} [t]
    \caption{%
        Performance improvements of applying a judge to assess evolution with (\cmark) and without (\xmark) using \ac{CoI}.
        The judge improves average performance ($\Delta S$) across both Differential Evolution (DE) and Genetic Algorithm (GA).
        With \ac{CoI}, DE sees a higher mean improvement (+0.87\% vs. +0.34\%) and a reduced worst-case drop in performance.
        GA also benefits, showing a larger average gain (+0.97\% vs. +0.65\%) and achieving the highest observed improvement overall (+3.20\%).
        These results highlight the effectiveness of using a judge, and especially in conjunction with \ac{CoI}, to guide the evolutionary process.
    }
    \label{tab:results-judge}
    \begin{tabularx}{\columnwidth}{ccRRR}
        \toprule
        Evolution              &          & \multicolumn{3}{c}{$\Delta S$}                               \\
        Algorithm              & \ac{CoI} & \textit{mean}                  & \textit{min} & \textit{max} \\
        \midrule
        \multirow[c]{2}{*}{DE} & \xmark   & +0.34\%                        & -0.86\%      & +2.80\%      \\
        \cline{2-5}
                               & \cmark   & \bfseries +0.87\%              & -0.18\%      & +2.58\%      \\
        \cline{1-5} \cline{2-5}
        \multirow[c]{2}{*}{GA} & \xmark   & +0.65\%                        & -0.22\%      & +1.79\%      \\
        \cline{2-5}
                               & \cmark   & \bfseries +0.97\%              & -0.35\%      & +3.20\%      \\
        \bottomrule
    \end{tabularx}
\end{table}


Similar to the enhancements achieved with \ac{CoI}, incorporating the judge in our approach consistently outperforms the baseline methods, as demonstrated in table \ref{tab:results-judge}, regardless of whether \ac{CoI} is utilized.
This result indicates that the judge can successfully detect and reject prompts which are determined to be of low quality before evaluation and therefore provides a positive answer to \rqn{3}.
The resulting increase in the number of high-quality prompts in the population before selection seems to yield an overall improvement in the performance of the final evolved prompt as observed in our results.

In combination, the results from tables \ref{tab:results-coi} and \ref{tab:results-judge} show -- in the ablation cases where either \ac{CoI} or the judge are removed -- that both concepts work best in combination, providing a positive answer to \rqn{2}.
This is to be expected, since the decomposition of the evolutionary prompt
also allows the judge to assess each smaller step separately, compared to judging the whole output including multiple steps over longer input-output pairs.

\begin{table}[h]
    \caption{%
        Relative score improvements of evolution strategies revised using human feedback (the subscript indicates the iteration) compared to DE and GA, respectively.
        Incorporating human feedback yields consistent performance gains across most tasks.
        On average, the second iteration of DE ($\text{DE}_2$) shows the highest mean improvement (+1.67\%) reflecting the notion of consecutive refinements, followed by the first DE refinement ($\text{DE}_1$, +1.11\%) and the GA refinement ($\text{GA}_1$, +0.75\%).
        The largest individual improvements are observed on TREC and ASSET, indicating that human feedback is particularly effective for tasks involving question classification and text simplification.
    }
    \label{tab:results-hf}
    \begin{tabularx}{\columnwidth}{lRRR}
        \toprule
        Task        & $\Delta S_{\text{DE}_1}$ & $\Delta S_{\text{DE}_2}$ & $\Delta S_{\text{GA}_1}$ \\
        \midrule
        \ac{AGNews} & +1.03\%                  & +1.54\%                  & +0.20\%                  \\
        \ac{ASSET}  & +1.74\%                  & +2.36\%                  & -0.18\%                  \\
        \ac{CR}     & +1.55\%                  & +2.10\%                  & +2.15\%                  \\
        \ac{MR}     & +0.15\%                  & +0.15\%                  & -0.05\%                  \\
        \ac{SAMSum} & -0.19\%                  & +1.09\%                  & +0.43\%                  \\
        \ac{SQuAD}  & +1.09\%                  & +0.95\%                  & +0.84\%                  \\
        \ac{SST} 2  & +0.90\%                  & +1.68\%                  & +0.88\%                  \\
        \ac{SST} 5  & +1.43\%                  & +2.06\%                  & +0.81\%                  \\
        Subj        & +0.20\%                  & +1.40\%                  & +0.20\%                  \\
        \ac{TREC}   & +3.20\%                  & +3.40\%                  & +2.20\%                  \\
        \midrule
        Mean        & \bfseries +1.11\%        & \bfseries +1.67\%        & \bfseries +0.75\%        \\
        \bottomrule
    \end{tabularx}
\end{table}

\subsection{\rqn{4}: Human Feedback}

The results presented in table \ref{tab:results-hf} demonstrate the relative performance improvements of evolution strategies incorporating human feedback ($\text{DE}_1$, $\text{DE}_2$, and $\text{GA}_1$) over their respective baseline methods (DE and GA) across multiple tasks.
Overall, $\text{DE}_2$ consistently outperforms its baseline, achieving the highest mean improvement of +1.67\%, compared to +1.11\% for $\text{DE}_1$.
$\text{GA}_1$ also benefits from human feedback, but shows a more modest mean improvement of +0.75\%.
While most tasks exhibit performance gains, there are a few instances where minimal or negative changes occur. 
Notably, the largest improvements are observed on the \ac{TREC} dataset showing substantial gains, with $\text{DE}_2$ achieving the highest relative improvement of +3.40\%.
However, there is a variance on individual tasks induced by the randomness of the evolution, but these findings suggest that integrating human feedback into evolution strategies can enhance performance, with DE-based approaches benefiting more noticeably than GA-based ones.\footnote{We note that the combination of judge-based and human feedback -- where humans corrected the model output if it was judged non-compliant with the instructions -- did not consistently enhance performance across all tasks. Consequently, we have opted not to present these results.}

Following this evaluation and to answer \rqn{4}, we can say that the evolutionary operator is effectively improved using human feedback as proposed in our approach.

\subsection{\rqn{5}: Effectiveness of different LLMs}

\begin{figure}[t!]
    \centering
    \includegraphics[width=\columnwidth]{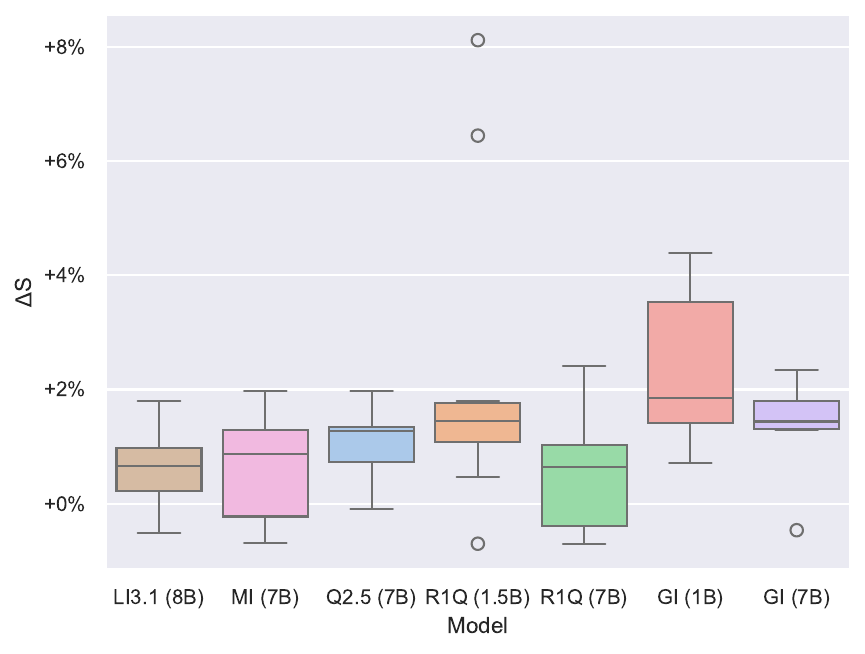}
    \caption{%
        Box plot illustrating the quantitative effectiveness of various \acp{LLM} based on performance metrics across our evaluation set.
        The models are listed in section \ref{sec:experimental-setup-other-models}.
        The y-axis represents the relative improvement in performance if \ac{CoI} and the judge are used.
        Mean performance is consistently improved across all tested models, with Gemma profiting the most.
    }
    \vspace{5mm}
    \label{fig:results-models}
\end{figure}

Figure \ref{fig:results-models} shows the performance difference between a baseline and runs using both the judge and \ac{CoI}, with different \acp{LLM} models.
The evaluation was performed for all tasks presented in \ref{sec:training-tasks}.
The results indicate that the proposed methods are effective across all tested models with an overall positive average improvement.
Interestingly, the smaller models, such as \texttt{R1Q (1.5B)} and \texttt{GI (1B)}, show a higher variance in performance, especially when compared directly to the larger variants of the same model (\texttt{R1Q (7B)} and \texttt{GI (7B)} respectively).

To answer \rqn{5}, we were able to demonstrate the effectiveness of our proposed method across a wide combination of tasks and models.

\begin{figure*}[t]
    \centering

    \begin{chatboxInput}
        \textbf{Input:}\\
        Step 1: Identify the different parts between Prompt 1 and Prompt 2:\\
        Prompt 1: Analyze the sentence and categorize it into one of five categories based on the sentiment: terrible, bad, okay, good, or great.\\
        Prompt 2:  Classify the given review into one of five categories: extremely negative (terrible), somewhat negative (bad), neutral (okay), somewhat positive (good), or extremely positive (great).
    \end{chatboxInput}

    \begin{chatboxResponse}
        \textbf{Response:}\\
        1. Identifying the different parts between Prompt 1 and Prompt 2:\\
        Prompt 1: Analyze the sentence and categorize it into one of five categories based on the sentiment: terrible, bad, okay, good, or great.\\
        Prompt 2:  Classify the given review into one of five categories: extremely negative (terrible), somewhat negative (bad), neutral (okay), somewhat positive (good), or extremely positive (great).\\
        Different parts:\\
        - "sentence" vs "review"\\
        - "analyze" vs "classify"\\
        \textcolor{red}{- "terrible, bad, okay, good, or great" vs "extremely negative (terrible), somewhat negative (bad), neutral (okay), somewhat positive (good), or extremely positive (great)"}\\
        \textcolor{red}{- "based on the sentiment" only appears in Prompt 1}\\
        \redsout{- "based on the sentiment" vs "based on the sentiment" (same phrase)}\\
        \redsout{- "five categories" vs "five categories" (same phrase)}\\
        \redsout{- The only difference is the subject of analysis, which is a sentence in Prompt 1 and a review in Prompt 2.}
    \end{chatboxResponse}

    \caption{An example for the first step of evolution for DE: The expected response mentions mutations for all spotted differences (marked in red) and omits the similarities as well as the last statement, which is evidently wrong (marked using strikeout in red). Demonstration samples for in-context learning are omitted for clarity.}
    \label{fig:evolution-sample-with-hf}
\end{figure*}

\section{Qualitative Analysis of Human Feedback}
\label{sec:analysis-human}
Figure \ref{fig:evolution-sample-with-hf} presents an example of human feedback within our approach.
Upon reviewing the model output from the first evolution step, it was observed that not all differences were identified, and some similarities were erroneously included.
Additionally, unexpected extraneous output was generated.
To address these issues, the prompt for the first evolution step was refined by incorporating the following instructions: "Output a list of all different parts and make sure that differences are only in the form of words and phrases." and "If the same phrase appears in both prompts, do not list it, i.e., do not list similarities."

This example also highlights the ambiguities associated with human involvement:
multiple instructions exist, just as there are various possible prompts for a task.
By iteratively analyzing the evolution model's output and refining the instructions accordingly, we can facilitate human feedback, ultimately enhancing the prompt optimization process.

Finally, we claim that inspecting the model output and adapting the instructions correspondingly can be accomplished in about half an hour, which is a reasonable time investment for the performance improvements achieved in our experiments.

\section{Conclusion}
We introduced and investigated extensions to evolutionary prompt optimization that leverage \ac{CoI}, an \ac{LLM}-based judge, human feedback and efficient evaluation methods to optimize prompts for a given task.

\ac{CoI}, by enabling greater control and better decision-making, can enhance performance in prompt optimization (\rqn{2}) and holds promise for broader applications.
In particular, when combined with judge-based assessment (\rqn{3}) and human feedback (\rqn{4}), it provides a robust framework for identifying optimal prompts in \ac{NLP} tasks.
Lastly, beyond performance improvements, reducing computational cost is also a key consideration.
Our efficient evaluation methods offer a significant reduction in computational overhead while maintaining performance during the search for optimal prompts (\rqn{1}).

We are convinced that our contributions, including investigations and releasing our code, help future research in the area, promoting the effective and efficient use of \acp{LLM} in \ac{NLP}, and especially help in grounding \acp{LLM} for better language understanding.

\section{Limitations}
In this work we only focus on the optimization of the prompts while not focusing on optimizing the verbalizer extracting the predictions for the tasks, which could be a potential improvement since the prompt can contain directives as to what to expect in the model output.

Furthermore, running multiple experiments on the same task can lead to different results due to the stochastic nature of \acp{LLM} and the evolutionary algorithms, providing a more reliable performance estimate.
However, since the experiments are time-intensive, we instead mitigate this effect by averaging the results over multiple tasks.
This also allows us to analyze the performance of our methods across a wide range of tasks, but may not be representative for individual tasks.

Also, although we optimize for faster runtimes and lower token usage, \acp{LLM} still require large amounts of compute resources and energy which potentially makes the methods and results presented in this paper inaccessible to some groups without access to such resources.
For example, a single optimization of a prompt for \ac{SAMSum} using the \textit{hardest first} strategy needed \texttt{4:24h} on a single NVIDIA A6000 GPU while the average GPU memory consumption was about 20GB.

\bibliography{references}

\end{document}